\documentclass[11pt,a4paper]{article}
\usepackage[hyperref]{acl2018}
\usepackage{times}
\usepackage{latexsym}
\usepackage{footnote}
\usepackage{tablefootnote}
\usepackage{enumitem}

\usepackage{bm}
\usepackage{multirow}
\usepackage{amsfonts}
\usepackage[boxed,linesnumbered,lined]{algorithm2e}
\usepackage{algpseudocode}
\usepackage{graphicx}
\usepackage{hyphenat}
\usepackage{verbatim}

\usepackage{CJKutf8}
\usepackage{graphicx}
\usepackage{nonfloat}
\usepackage{fmtcount}
\usepackage{threeparttable}
\usepackage{caption}

\usepackage{url}

\aclfinalcopy 

\newcommand{\dataset}{CAIL2018 }

\title{CAIL2018: A Large-Scale Legal Dataset for Judgment Prediction}

\author{Chaojun Xiao$^{1\ast}$ Haoxi Zhong$^{1}$\thanks{$ $ $ $ $ $ indicates equal contribution.} $ $ Zhipeng Guo$^{1}$ Cunchao Tu$^{1}$ Zhiyuan Liu$^{1}$ \\ \textbf{Maosong Sun$^{1}$ Yansong Feng$^{2}$ Xianpei Han$^{3}$ Zhen Hu$^{4}$ Heng Wang$^4$ Jianfeng Xu$^{5}$}\\
\normalsize $^{1}$Department of Computer Science and Technology, Tsinghua University, China \\
\normalsize $^{2}$Institute of Computer Science and Technology, Peking University, China\\
\normalsize $^{3}$Institute of Software, Chinese Academy of Sciences, China\\
\normalsize $^{4}$China Justice Big Data Institute\\
\normalsize $^{5}$Supreme People’s Court, China\\
}

\date{}

\begin{document}
\begin{CJK*}{UTF8}{gbsn}

\maketitle
\begin{abstract}
In this paper, we introduce the \textbf{C}hinese \textbf{AI} and \textbf{L}aw challenge dataset (CAIL2018),  the first large-scale Chinese legal dataset for judgment prediction. \dataset contains more than $2.6$ million criminal cases published by the Supreme People's Court of China, which are several times larger than other datasets in existing works on judgment prediction. Moreover, the annotations of judgment results are more detailed and rich. It consists of applicable law articles, charges, and prison terms, which are expected to be inferred according to the fact descriptions of cases. For comparison, we implement several conventional text classification baselines for judgment prediction and experimental results show that it is still a challenge for current models to predict the judgment results of legal cases, especially on prison terms. To help the researchers make improvements on legal judgment prediction, both \dataset and baselines will be released after the CAIL competition\footnote{http://cail.cipsc.org.cn/}.

\end{abstract}

\section{Introduction} \label{section:intro}
The task of Legal Judgment Prediction(LJP) aims to empower machine to predict the judgment results of legal cases after reading fact descriptions. It has been studied for decades. Due to the limitation of publicly available cases, early works~\cite{lauderdale2012supreme,segal1984predicting,keown1980mathematical,ulmer1963quantitative,nagel1963applying,kort1957predicting} usually conduct statistical analysis on the judgment results over a small number of cases rather than predicting them. With the development of machine learning algorithms, some works take LJP as a text classification task and propose to extract efficient features from fact descriptons~\cite{Liu2017A,Sulea2017Predicting,aletras2016predicting,lin2012,liu2006exploring}. These works are still restricted to particular case types and suffer from generalization issue when applied to other scenarios.

Inspired by the success of deep learning techniques on natural language processing tasks, researchers attempt to employ neural models to handle judgment prediction task under the text classification framework~\cite{luo2017learning,hu2018few}. However, there is not a publicly accessible high-quality dataset for LJP yet. Therefore, we collect and release the first large-scale dataset for LJP, i.e., CAIL2018, to encourage further explorations on this task and other advanced legal intelligence algorithms.

\dataset consists of more than $2.6$ million criminal cases, which are collected from \url{http://wenshu.court.gov.cn/} published by the Supreme People's Court of China. These documents serve as the reference for professionals to improve their working efficiency and are expected to benefit researches on legal intelligent systems. 

Specifically, each case in \dataset consists of two parts, i.e., fact description and corresponding judgment result. Here, the judgment result of each case is refined into $3$ representative ones, including relevant law articles, charges, and prison terms. Comparing with other datasets used by existing LJP works, \dataset is on a larger scale and reserves richer annotations of judgment results. Totally, \dataset contains $2,676,075$ criminal cases, which are annotated with $183$ criminal law articles and $202$ criminal charges. Both the number of cases and the number of labels are several times than other closed-source LJP datasets.

In the following parts, we give a detailed introduction to the construction of \dataset and the LJP results of baseline methods on this dataset.

\begin{table*}[!t]\centering
\small
\begin{tabular}{c|c|c|c|c}
\hline
\textbf{Fact} & \textbf{Relevant Law Article} & \textbf{Charge} & \textbf{Prison Term} & \textbf {Defendant}\\ 
\hline
被告人胡某... & 刑法第234条 & 故意伤害 & 12个月 & 胡某 \\
The Defendant Hu... & 234th article of criminal law & intentional injury & 12 months & Miss./Mr. Hu\\

\hline
\end{tabular}
\caption{An example in \dataset.}
\label{table:example}
\end{table*}

\section{Dataset Construction}
\label{section:CAIL_dataset}

We construct \dataset from $5,730,302$ criminal documents collected from China Judgments Online\footnote{http://wenshu.court.gov.cn/}. There documents of criminal cases belong to five types, including \emph{judgment}, \emph{verdict}, \emph{conciliation statement}, \emph{decision letter}, and \emph{notice}. For LJP, we only concern on these cases with judgment results. Therefore, we only keep these judgment documents for training LJP models.

Each original document is well-structured and divided into several parts, e.g., fact description, court view, parties, judgment result and other information. Therefore, we take the fact part as input and extract applicable law articles, charges and prison terms from referee result with regular expressions.

Since many criminal cases own multiple defendants, which would increase the difficulty of LJP greatly, we only retain the cases with a single defendant. 

In addition, there are also many low-frequency charges(e.g. insult the national flag, jailbreak) and law articles. We filter out cases with those charges and law articles whose frequency is smaller than $30$. Besides, the top $102$ law articles in Chinese Criminal Law are not relevant to specific charges, we filter out these law articles and charges as well.

After preprocessing, the dataset contains $2,676,075$ criminal cases, $183$ criminal law articles, $202$ charges and prison term.  We also show an instance in \dataset in Table~\ref{table:example}.

It is worth noting that, the distribution of different categories in \dataset is quite imbalanced. Considering the number of various charges, the top $10$ charges cover $79.0\%$ cases. On the contrary, the bottom $10$ charges only cover $0.12\%$ cases. The imbalance issue in \dataset makes it challenging to predict low-frequency charges and law articles.

\section{Experiments}

In this section, we implement and evaluate several typical text classification baselines on three subtasks of LJP, including law articles, charges, and prison terms.

\begin{table*}[!h]
\centering
\begin{tabular}{c|ccc|ccc|ccc}
\hline
\textbf{Tasks}  & \multicolumn{3}{c|}{Charges} & \multicolumn{3}{c|}{Relevant Articles} & \multicolumn{3}{c}{Terms of Penalty} \\ \hline
\textbf{Metrics}            & Acc.\% & MP\% & MR\% & Acc.\% & MP\% & MR\% & Acc.\% & MP\% & MR\% \\ \hline
\textbf{FastText}             & 94.3   & 50.9 & 39.7 & 93.3   & 45.8 & 38.1 & 74.6   & 48.0 & 24.5 \\ \hline
\textbf{TFIDF+SVM}             & 94.0   & 73.9 & 56.2 & 92.9   & 71.8 & 52.4 & 75.4   & 75.4 & 46.1 \\ \hline
\textbf{CNN}                  & 97.6    & 37.0 & 21.4 & 97.6 & 37.4 & 21.8 & 78.2 & 45.5 & 36.1   \\ \hline
\end{tabular}
\caption{LJP results on CAIL.}
\label{table:results}
\end{table*}

\subsection{Baselines}
We select following $3$ baselines for comparison:

\textbf{TFIDF+ SVM:} Term-frequency inverse document frequency (TFIDF)~\cite{salton1988term} is an efficient method to extract word features and Support Vector Machine (SVM)~\cite{suykens1999least} is a representative classification model. We implement TFIDF to extract text features and employ SVM with linear kernel to train the classifier.

\textbf{FastText:} FastText~\cite{joulin2017bag} is a simple and efficient approach for text classification based on N-grams and Hierarchical softmax~\cite{mikolov2013distributed}.

\textbf{CNN:} Convolutional Neural Network(CNN) has been proven efficient in text classification~\cite{kim2014convolutional}. We employ the CNN with multiple filters to encode fact descriptions.


\subsection{Implementation Details}
For all the methods, we randomly select $1,710,856$ cases for training and $965,219$ cases for testing. Since all fact descriptions are written in Chinese, we employ THULAC~\cite{sun2016thulac} for word segmentation. For TFIDF+SVM model, we limit the feature size to $5,000$. For neural-based model, we employ Skip-Gram model~\cite{mikolov2013distributed} to train word embeddings with $200$ dimensions.

For CNN, we set the maximum length of a case description to $4,096$, the filter widths to $(2,3,4,5)$ with each filter size to $64$ for consistency. 

For training, we employ Adam~\cite{kingma2014adam} as the optimizer. We set the learning rate to $0.001$, the dropout rate to $0.5$, and the batch size to $128$.

\subsection{Results and Analysis}
We evaluate baseline models with several metrics, including accuracy(Acc.), macro-precision(MP) and macro-recall(MR) which are widely used in the classification task. Experimental results on the test set are shown in Table~\ref{table:results}.

From this table, we find that current models can achieve considerable results on the accuracy of charges prediction and relevant law articles prediction. However, the results of MP and MR show that LJP is still a huge challenge due to the lack of training data and imbalance issue.

\section{Conclusion}
In this work, we release the first large-scale legal judgment prediction dataset, CAIL2018. Comparing with existing LJP datasets, \dataset is the largest LJP dataset so far and publicly available. Moreover, \dataset reserves more detailed annotations, which is consistent with real-world scenarios. Experiments demonstrate that LJP is still challenging and leave plenty of room to make improvements.

\bibliography{reference}

\begin{thebibliography}{}
\expandafter\ifx\csname natexlab\endcsname\relax\def\natexlab#1{#1}\fi

\bibitem[{Aletras et~al.(2016)Aletras, Tsarapatsanis, Preotiuc-Pietro, and
  Lampos}]{aletras2016predicting}
Nikolaos Aletras, Dimitrios Tsarapatsanis, Daniel Preotiuc-Pietro, and
  Vasileios Lampos. 2016.
\newblock Predicting judicial decisions of the european court of human rights:
  A natural language processing perspective.
\newblock {\em PeerJ Computer Science\/} 2.

\bibitem[{Hu et~al.(2018)Hu, Li, Tu, Liu, and Sun}]{hu2018few}
Zikun Hu, Xiang Li, Cunchao Tu, Zhiyuan Liu, and Maosong Sun. 2018.
\newblock Few-shot charge prediction with discriminative legal attributes.
\newblock In {\em Proceedings of COLING\/}.

\bibitem[{Joulin et~al.(2017)Joulin, Grave, Bojanowski, and
  Mikolov}]{joulin2017bag}
Armand Joulin, Edouard Grave, Piotr Bojanowski, and Tomas Mikolov. 2017.
\newblock Bag of tricks for efficient text classification.
\newblock In {\em Proceedings of EACL\/}.

\bibitem[{Keown(1980)}]{keown1980mathematical}
R~Keown. 1980.
\newblock Mathematical models for legal prediction.
\newblock {\em Computer/LJ\/} 2:829.

\bibitem[{Kim(2014)}]{kim2014convolutional}
Yoon Kim. 2014.
\newblock Convolutional neural networks for sentence classification.
\newblock In {\em Proceedings of EMNLP\/}.

\bibitem[{Kingma and Ba(2015)}]{kingma2014adam}
Diederik Kingma and Jimmy Ba. 2015.
\newblock Adam: A method for stochastic optimization.
\newblock In {\em Proceedings of ICLR\/}.

\bibitem[{Kort(1957)}]{kort1957predicting}
Fred Kort. 1957.
\newblock Predicting supreme court decisions mathematically: A quantitative
  analysis of the "right to counsel" cases.
\newblock {\em American Political Science Review\/} 51(1):1--12.

\bibitem[{Lauderdale and Clark(2012)}]{lauderdale2012supreme}
Benjamin~E Lauderdale and Tom~S Clark. 2012.
\newblock The supreme court's many median justices.
\newblock {\em American Political Science Review\/} 106(4):847--866.

\bibitem[{Lin et~al.(2012)Lin, Kuo, Chang, Yen, Chen, and Lin}]{lin2012}
Wan-Chen Lin, Tsung-Ting Kuo, Tung-Jia Chang, Chueh-An Yen, Chao-Ju Chen, and
  Shou-de Lin. 2012.
\newblock Exploiting machine learning models for chinese legal documents
  labeling, case classification, and sentencing prediction.
\newblock In {\em Processdings of ROCLING\/}. page 140.

\bibitem[{Liu and Hsieh(2006)}]{liu2006exploring}
Chao-Lin Liu and Chwen-Dar Hsieh. 2006.
\newblock Exploring phrase-based classification of judicial documents for
  criminal charges in chinese.
\newblock In {\em Proceedings of ISMIS\/}. pages 681--690.

\bibitem[{Liu and Chen(2017)}]{Liu2017A}
Yi~Hung Liu and Yen~Liang Chen. 2017.
\newblock A two-phase sentiment analysis approach for judgement prediction.
\newblock {\em Journal of Information Science\/} .

\bibitem[{Luo et~al.(2017)Luo, Feng, Xu, Zhang, and Zhao}]{luo2017learning}
Bingfeng Luo, Yansong Feng, Jianbo Xu, Xiang Zhang, and Dongyan Zhao. 2017.
\newblock Learning to predict charges for criminal cases with legal basis.
\newblock In {\em Proceedings of EMNLP\/}.

\bibitem[{Mikolov et~al.(2013)Mikolov, Sutskever, Chen, Corrado, and
  Dean}]{mikolov2013distributed}
Tomas Mikolov, Ilya Sutskever, Kai Chen, Greg~S Corrado, and Jeff Dean. 2013.
\newblock Distributed representations of words and phrases and their
  compositionality.
\newblock In {\em Proceedings of NIPS\/}. pages 3111--3119.

\bibitem[{Nagel(1963)}]{nagel1963applying}
Stuart~S Nagel. 1963.
\newblock Applying correlation analysis to case prediction.
\newblock {\em Tex. L. Rev.\/} 42:1006.

\bibitem[{Salton and Buckley(1988)}]{salton1988term}
Gerard Salton and Christopher Buckley. 1988.
\newblock Term-weighting approaches in automatic text retrieval.
\newblock {\em Information processing \& management\/} 24(5):513--523.

\bibitem[{Segal(1984)}]{segal1984predicting}
Jeffrey~A Segal. 1984.
\newblock Predicting supreme court cases probabilistically: The search and
  seizure cases, 1962-1981.
\newblock {\em American Political Science Review\/} 78(4):891--900.

\bibitem[{Sulea et~al.(2017)Sulea, Zampieri, Vela, and
  Genabith}]{Sulea2017Predicting}
Octavia~Maria Sulea, Marcos Zampieri, Mihaela Vela, and Josef~Van Genabith.
  2017.
\newblock Exploring the use of text classi cation in the legal domain.
\newblock In {\em Proceedings of ASAIL workshop\/}.

\bibitem[{Sun et~al.(2016)Sun, Chen, Zhang, Guo, and Liu}]{sun2016thulac}
Maosong Sun, Xinxiong Chen, Kaixu Zhang, Zhipeng Guo, and Zhiyuan Liu. 2016.
\newblock Thulac: An efficient lexical analyzer for chinese. .

\bibitem[{Suykens and Vandewalle(1999)}]{suykens1999least}
Johan~AK Suykens and Joos Vandewalle. 1999.
\newblock Least squares support vector machine classifiers.
\newblock {\em Neural processing letters\/} 9(3):293--300.

\bibitem[{Tang et~al.(2015)Tang, Qin, and Liu}]{tang2015document}
Duyu Tang, Bing Qin, and Ting Liu. 2015.
\newblock Document modeling with gated recurrent neural network for sentiment
  classification.
\newblock In {\em Proceedings of EMNLP\/}. pages 1422--1432.

\bibitem[{Ulmer(1963)}]{ulmer1963quantitative}
S~Sidney Ulmer. 1963.
\newblock Quantitative analysis of judicial processes: Some practical and
  theoretical applications.
\newblock {\em Law \& Contemp. Probs.\/} 28:164.

\end{thebibliography}
\bibliographystyle{acl_natbib}

\end{CJK*}

\end{document}